\documentclass[runningheads]{llncs}
\usepackage{graphicx}
\usepackage{microtype}
\usepackage{booktabs}
\usepackage{makecell}
\usepackage{multirow}
\usepackage{amsmath}
\usepackage{amssymb}
\usepackage{marvosym}
\usepackage[table,xcdraw]{xcolor}
\newcommand{\mycolor}[1]{\textcolor{blue}{#1}}

\def\ie{\emph{i.e.}}

\begin{document}

\title{Attention-Guided Perturbation Network for Industrial Anomaly Detection}

\titlerunning{AGPNet}

\author{ 
Tingfeng Huang\inst{1}\textsuperscript{$\dagger$} \and
Weijia Kong\inst{1}\textsuperscript{$\dagger$} \and
Huan Liu\inst{2} \and
Shengjie Chen\inst{2}\textsuperscript{(\Letter)} \and
Bao Zhou\inst{2}\textsuperscript{(\Letter)} \and
Ligang Zhang\inst{2} \and 
Xinwei He\inst{1}  
}

\authorrunning{T. Huang et al.}

\institute{
Huazhong Agricultural University, China\\
\email{\{kst0D83,wjkong\}@webmail.hzau.edu.cn}
\and
Henan Sanrui Technology Co., Ltd., China\\
\email{740393637@qq.com, zhb1220@126.com}\\
}

\maketitle

\begingroup
\renewcommand{\thefootnote}{}
\footnotetext{$\dagger$ Equal contribution.
$^{(\textrm{\Letter})}$ Corresponding author.}
\endgroup

\begin{abstract}
In unsupervised image anomaly detection, reconstruction-based methods learn normal patterns for data reconstruction, but often undesirably reconstruct anomalous regions at inference, resulting in missed detections.
To alleviate this, existing approaches perturb normal samples in a sample-agnostic manner by uniformly injecting noise, ignoring that foreground regions are more critical for robust reconstruction.
To address this limitation, we propose Attention-Guided Perturbation Network (AGPNet), a novel reconstruction framework for industrial anomaly detection. AGPNet introduces sample-aware attention masks to guide perturbations, enhancing the learning of invariant normal patterns at important locations.
AGPNet consists of two branches, a reconstruction branch and an auxiliary attention-based perturbation one.
The reconstruction branch focuses on learning to reconstruct normal samples, while the auxiliary branch generates attention masks to guide noise perturbation.
By applying stronger perturbations to salient regions, the reconstruction branch learns intrinsic normal patterns in a more comprehensive and robust manner.
Extensive experiments on MVTec-AD, VisA, and MVTec-3D show that AGPNet achieves competitive or leading performance across few-shot, one-class, and multi-class settings, with particularly strong performance in few-shot anomaly localization.

\end{abstract}

\section{Introduction}
\label{sec:intro}
Automatically detecting and localizing anomalies in industrial images has extensive applications in manufacturing~\cite{bergmann2019mvtec,defard2021padim}. However, this task typically encounters the challenge of \emph{cold-start}~\cite{zavrtanik2021draem} where gathering normal samples is easy, whereas acquiring anomalous samples is costly or infeasible. 
In recent years, unsupervised anomaly detection (UAD) methods that model normality using only normal data have been extensively studied~\cite{zhang2023destseg,RD4AD}, among which reconstruction-based approaches stand out for their simplicity and scalability.

Reconstruction methods rely on one core assumption: \textbf{reconstruction networks exposed to normal data alone can reconstruct normal ones well, but find difficulty in reconstructing abnormal ones}.
Thus, an anomaly map can be computed by analyzing the reconstruction error of the input.
As per this assumption, some methods attempt to reconstruct input images with popular generative models~\cite{yan2021learning,he2024diad}.
However, pixel-level reconstruction often becomes ineffective when normal and anomalous regions share similar appearances~\cite{you2022adtr}. 
Recent studies shift towards feature-level reconstruction, which provides richer contextual semantics with large-scale pretrained models~\cite{you2022unified,deng2009imagenet}. 

\begin{figure}
\includegraphics[width=\textwidth]{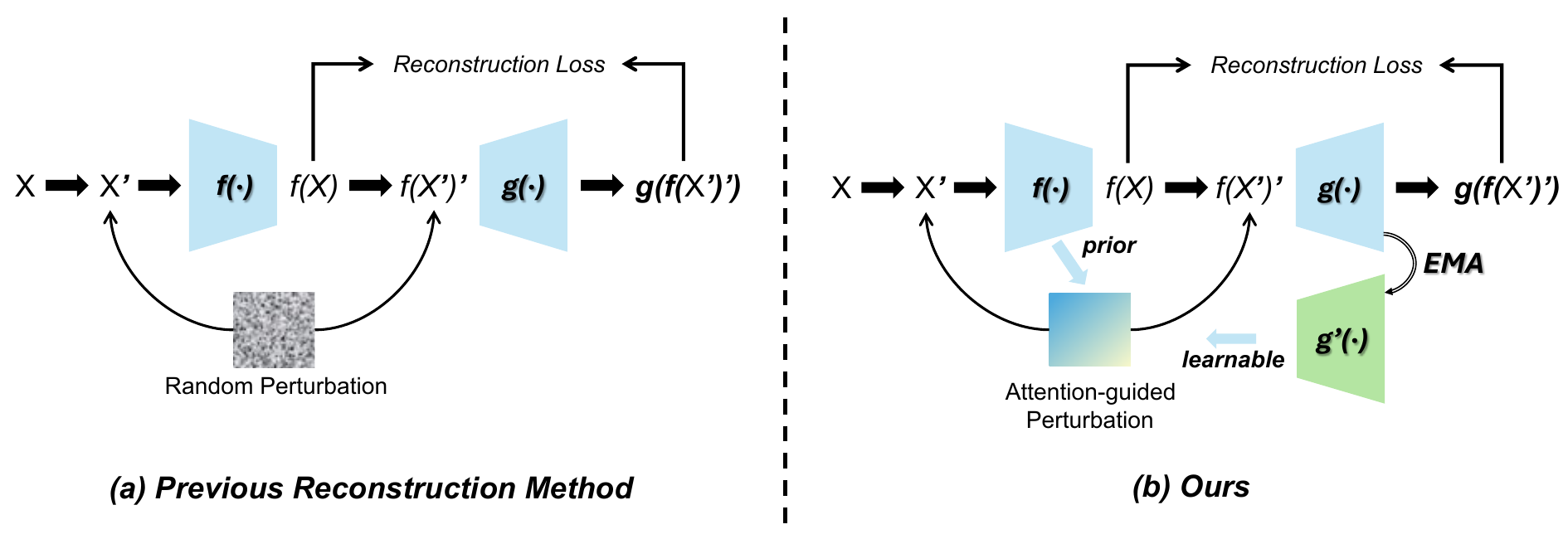}
\caption{Comparisons between ours and existing perturbation strategies in reconstruction paradigm for anomaly detection. Compared with (a) existing networks adopting fixed or random perturbations, (b) we introduce to learn to guide the perturbation with sample-aware masks from the network and help to learn a better reconstruction model.} \label{differ}
\vspace{-2mm}
\end{figure}

Despite remarkable progress, \textbf{this assumption behind reconstruction methods, however, does not always hold in practice}, as deep neural networks may generalize to anomalous regions and reconstruct them unpredictably. 
To mitigate this, prior works introduce input perturbations and map the perturbed samples back to normal ones to suppress anomaly reconstruction (Figure~\ref{differ} (a))~\cite{yan2021learning,zavrtanik2021draem,you2022unified}. 
By perturbing the input, the reconstruction network is encouraged to learn invariant normal patterns. 
However, existing approaches typically rely on fixed or random perturbations applied indiscriminately to input images or features, without considering their contextual or semantic characteristics.
To address this problem, we propose to leverage \textbf{sample-specific content and structural priors for perturbation}. 
To this end, we introduce a simple yet effective reconstruction framework, \textbf{AGPNet} (Figure~\ref{fig:framework}), for unsupervised anomaly detection. 
The core idea is to perturb normal inputs according to the importance of each spatial location during training, guiding the reconstruction network to focus on critical areas and learn \emph{invariant normal patterns} more effectively.
To achieve this goal, we design an auxiliary branch to generate attention masks for the noise-adding process (Figure~\ref{differ} (b)), enabling perturbations to be applied in a sample-aware manner. 
Specifically, the auxiliary branch integrates two complementary cues to derive the final attention masks. 
The first cue comes from attention maps of pre-trained feature extraction layers, which provide strong priors on foreground regions and important localization information for reconstruction.  
The second cue is obtained from the momentum-distilled~\cite{tarvainen2018mean} attention maps of the decoder layers in the main branch, reflecting the importance of each location for the reconstruction task.
By incorporating these cues, the auxiliary branch dynamically emulates the reconstruction behavior and perturbs important regions more aggressively as training progresses. 
This strategy explicitly exposes the model to harder reconstruction cases, effectively synthesizing challenging anomalous patterns during training and forcing the reconstruction network to learn more discriminative and compact normality boundaries. 
Finally, the normal input is perturbed with Gaussian noise weighted by the attention masks, which further encourages the framework to reason about invariant normal patterns at both low and high levels.
After training, the auxiliary branch is removed.
Extensive experiments validate the effectiveness of AGPNet. 

Our contributions are summarized as follows:
\begin{itemize}
\item We propose AGPNet, a concise reconstruction-based framework that introduces a sample-aware perturbation strategy. 
Unlike prior methods that apply fixed or random perturbations uniformly across spatial locations, AGPNet selectively perturbs critical regions for each sample to facilitate more discriminative reconstruction.
\item We design an attention-guided perturbation mechanism that integrates both pretrained feature priors and learnable reconstruction-aware cues, enabling the model to capture invariant normal patterns across diverse categories.
\item Extensive experiments demonstrate competitive or leading performance in multi-class and one-class settings, strong few-shot anomaly localization, and a favorable accuracy--efficiency trade-off.
\end{itemize}

\section{Related Work}
Existing UAD algorithms can be broadly divided into three groups: synthesizing-based, embedding-based, and reconstruction-based methods. 

\textbf{Synthesizing-based methods}~\cite{cutpaste,zavrtanik2021draem} synthesize anomalies for training. For instance, DRAEM~\cite{zavrtanik2021draem} attempts to blend predefined texture images with normal samples. 
However, due to the uncertainty and unpredictability of anomalies, it is impossible to synthesize all types of real anomalies with high fidelity.
\textbf{Embedding-based methods}~\cite{defard2021padim,roth2022towards} typically work by first utilizing pretrained models to embed normal images and then employing statistical models to model the normal distributions. During testing, samples are regarded as being abnormal if they are far away from the normal distributions. For instance, PaDiM~\cite{defard2021padim} models embedded patch features with multivariate Gaussian distribution. PatchCore~\cite{roth2022towards} exploits memory banks to store nominal features. However, they generally demand more computational resources to store the normal embeddings for the training set.
\textbf{Reconstruction-based methods}~\cite{zavrtanik2021reconstruction,zhou2020encoding} assume that models trained solely on normal data can faithfully reconstruct normal samples while failing on anomalous ones during inference. 
However, this assumption is often violated due to the strong and uncontrollable generalization of deep neural networks.
Therefore, many strategies have been proposed to address this issue. 
Some methods try to add prior from the images~\cite{zhou2020encoding} as guidance for reconstructions. For instance, P-Net~\cite{zhou2020encoding} proposes to feed the structure features into the reconstruction network.
Other methods~\cite{ye2020attribute,liu2023fair} frame anomaly detection as the image restoration problem by corrupting selected content in images and then learning to restore the corruptions. 
Several methods~\cite{li2020superpixel,zavrtanik2021reconstruction} take the idea of image inpainting by first masking the image randomly and then learning to recover it.
Recent advances continue to demonstrate the superiority of ViT-based architectures in anomaly detection, including pure vision transformer designs for zero-shot tasks (e.g., VisualAD~\cite{VisualAD}). For instance, ViTAD~\cite{zhang2025exploring} achieves state-of-the-art performance with a plain transformer decoder, yet it still relies on uniform random perturbations that fail to prioritize semantically critical regions.

Despite promising results, the above methods focus on the \emph{``one-for-one''} paradigm, which trains one separate model for each category and thus is more computationally expensive. 

Recently, the more challenging \emph{``one-for-all''} paradigm~\cite{you2022unified,meng2024moead} has attracted increasing attention, which aims to train a single model across all categories. 
Following this direction, unified frameworks like UniNet~\cite{UniNet}, Dinomaly~\cite{Dinomaly}, and INP-Former~\cite{INP-Former} further push the performance boundary of multi-class anomaly detection. However, these methods typically rely on large ViT-B backbones and high-resolution inputs, leading to significantly higher computational costs and inference latency.
Furthermore, to cope with complex real-world manufacturing, the latest literature has witnessed a surge in zero-shot learning frameworks guided by vision-language models (e.g., GS-CLIP~\cite{GS_CLIP}) and the exploration of multimodal 3D anomaly detection (e.g., Real-IAD D³~\cite{RealIAD_D3}). Despite these broader trends, efficiently modeling normality in the 2D unified setting remains a fundamental and challenging prerequisite.

Our method is reconstruction-based and applicable to both paradigms. 
Different from prior approaches that rely on sophisticated architectural designs, we introduce an attention-guided perturbation scheme that enables a simple ViT-based decoder to learn invariant normal features. 
By leveraging model-derived attention masks to guide perturbation, our reconstruction network can have a better understanding of the intrinsic and important local normal patterns, which help suppress the anomaly reconstructions effectively. 

\section{AGPNet}

\begin{figure}[t]
\centering 
\includegraphics[width=\textwidth]{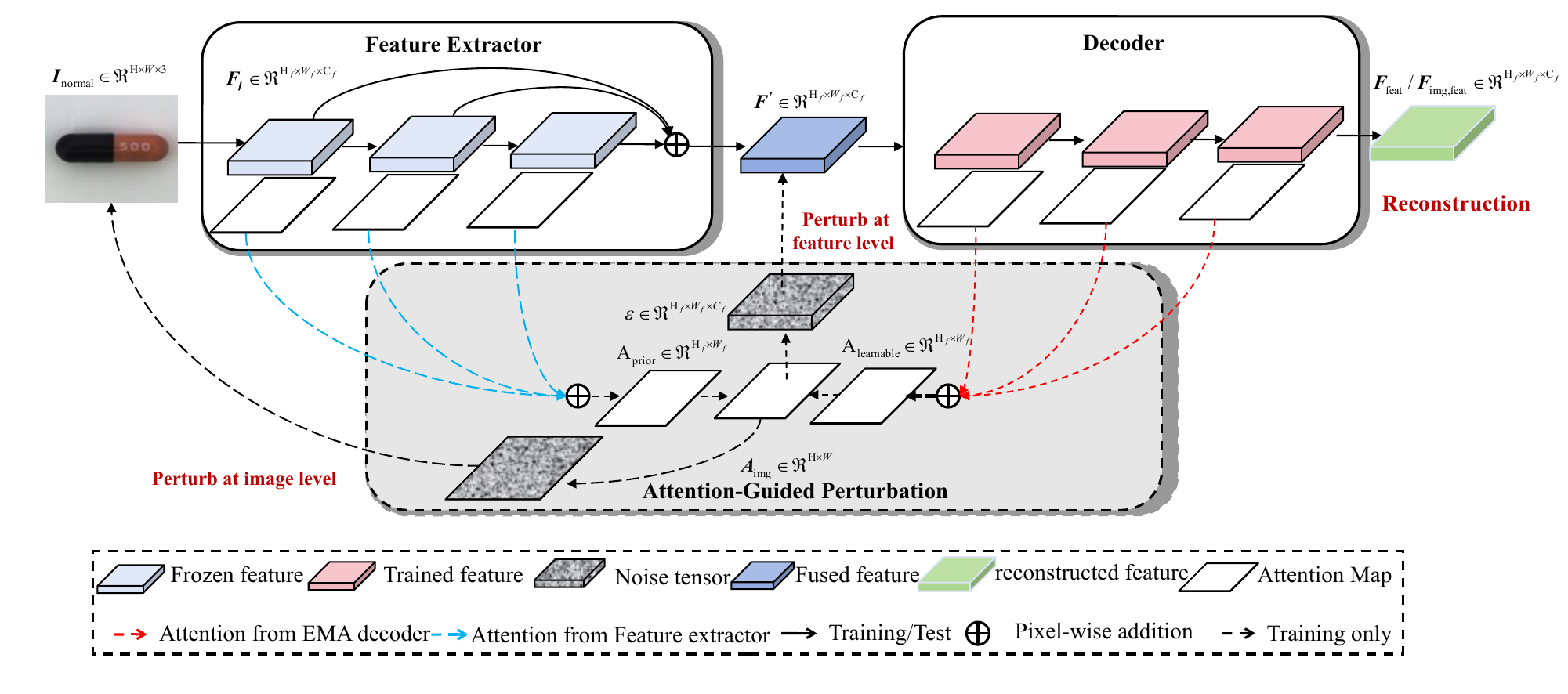}
\caption{
        Overview of AGPNet.
        During training, given an input normal image $I_{normal}$, the main reconstruction branch is used for reconstruction, while the perturbation branch aims to generate attention masks based on the main branch for perturbation at both image and feature levels, making the reconstruction network focus on the important local details. 
    }\label{fig:framework}
    \vspace{-2mm}
\end{figure}

As shown in Figure~\ref{fig:framework}, it consists of two branches: the main reconstruction branch aims to reconstruct the inputs, while the auxiliary attention-guided perturbation branch generates attention masks to guide the perturbation for reconstruction. 
\subsection{Main Reconstruction Branch}

\noindent\textbf{Feature Extractor.} Given an input RGB image $I\in \mathbb{R}^{H \times W \times3}$,
we feed it into a pretrained network $\phi(\cdot)$ and only take a subset of $L$ layer outputs. 
Any off-the-shelf pre-trained ViT-like model, which relies on attention mechanisms, can readily serve as the feature extractor.
In this paper, we follow previous work~\cite{reiss2022anomaly} and adopt DINO~\cite{caron2021emerging} as our backbone for its superior representation capacity.
We denote $F_l \in \mathbb{R}^{H_f \times W_f \times C_f}$ as the $l$-th output, and $\sigma(\cdot)$ as the layer normalization function.
The representations from specific layers will undergo layer normalization individually, followed by summation to derive clean feature $F_{\text{clean}} \in \mathbb{R}^{H_f \times W_f \times C_f}$:
\begin{equation}
    F_{\text{clean}} = \sum_{l=1}^{L}{\sigma(F_l)},
\end{equation}
which will serve as the reconstruction target and be perturbed for further processing by the decoder. 

Meanwhile, in addition to the $L$ outputs, we propose to reuse the associated attention maps in DINO~\cite{caron2021emerging} to guide the decoder training with \emph{minimal efforts}. 
Those attention maps, denoted by \textbf{\textbf{$A_l \in \mathbb{R}^{H_l \times W_l}$}}, can provide the important fine-grained prior cues for the perturbation branch (Sec.~\ref{subsec:agp}). 

\noindent\textbf{Decoder.} The decoder aims to decode the pretrained features of \emph{normal} data in the training set. 
In our framework, we simply adopt a plain vision-transformer-based decoder thanks to an auxiliary attention-based perturbation branch.
Note that our decoder is lightweight and has only four layers, yet it demonstrates strong empirical performance. 

\subsection{Attention-guided Perturbation Branch}
~\label{subsec:agp}

\noindent\textbf{Attention Mask Generation.}
The goal of the auxiliary branch is to perturb normal input for training under the guidance of attention masks. 
To calculate the attention mask, we leverage the attention maps from the feature extractor and the momentum distillation of the decoder, which is formulated as below: 
\begin{equation}
\begin{array}{rl}
    A_{\text{final}} =&\Phi_{\text{norm}}(A_{\text{prior}}) +\Phi_{\text{norm}} (A_\text{learn})        \\
   \textrm{where}  & \\
   & \left\{
   \begin{array}  {l}
   \Phi_{\text{norm}} := \textrm{max-min normalization function.}\\
   A_{\text{prior}} = \Phi_{\text{aggr}}(\{A_l\}_{l=1}^L) \\
   A_{\text{learn}} = \Phi_{\text{aggr}}(\{A_k\}_{k=1}^K)\\
   \end{array}
   \right.
\end{array}
\end{equation}
The Spatial max-min normalization to $[0, 1]$ is applied for $\Phi_{\text{norm}}$. This bridges the scale discrepancy between $A_{\text{prior}}$ and $A_{\text{learn}}$ for a balanced and stable fusion. For aggregation of multi-layer attention maps from either the feature extractor or the decoder, we sequentially apply average pooling across the layers, average the attention weights along the query token dimension, sum across the multiple attention heads and reshape matching the feature resolution. Replacing average- with max- pooling yields negligible differences ($<$0.1\%).

As discussed earlier, the attention weights from the feature extractor directly reflect the importance of each localization in the feature maps~\cite{zeiler2014visualizing}. Therefore, based on the extracted subset of attention weights $\{A_l\}_{l=1}^L$, we compute the prior attention mask by 
$   A_{\text{prior}} = \Phi_{\text{aggr}}(\{A_l\}_{l=1}^L)$, 
where $\Phi_{aggr}$ denotes the aggregation operation. In practice, here we simply use element-wise average pooling. 

Second, the attention weights within the decoder are learned to aggregate important contextual cues for reconstruction. We further propose to integrate them. 
However, these attention weights fluctuate rapidly with high variance, especially at the beginning of the training. For stabilization, we employ mean-distillation~\cite{tarvainen2018mean}, which includes an exponential-moving-average (EMA) version of the decoder as the teacher. 
\begin{equation}
     \theta_{\text{{md}}} = \eta \cdot \theta_{\text{{md}}} + (1 - \eta) \cdot \theta_{\text{{dec}}}
 \end{equation}
where $\theta_{\text{{md}}}$ and $\theta_{\text{{dec}}}$ denote the parameters of the teacher and student (\ie, the decoder) models, respectively. $\eta$ controls the weight assigned to previous teacher parameters.
The self-attention weights $\{A_k\}_{k=1}^K$ from the teacher are taken to compute the learnable attention masks by
 $  A_{\text{learn}} = \Phi_{\text{aggr}}(\{A_k\}_{k=1}^K)$.
Lastly, we derive the final mask as $A_{\text{final}} =\Phi_{\text{norm}}(A_{\text{prior}}) +\Phi_{\text{norm}} (A_\text{learn})$
which will be utilized to guide the following noise-adding process. 

\noindent\textbf{Attention-Guided Noise.}
Synthesizing pseudo-anomalies is pivotal for learning a compact decision boundary~\cite{zavrtanik2021draem,cutpaste}. However, since simulating the infinite variety of real anomalies is intractable, we argue that selectively perturbing critical locations in the feature space is a more efficient and robust strategy. By corrupting these informative regions, the model is compelled to reason about intrinsic semantic relations.
To this end, we simply add Gaussian noise weighted by the attention mask in the normal feature space to synthesize the hard abnormal samples. Specifically, a noise tensor $\mathcal{E} \in \mathbb{R}^{H_f \times \ W_f \times C_f}$ is first generated with each entry simply drawn from an \textbf{i.i.d} Gaussian distribution $\mathcal{N}(\mu, \sigma)$. Then we add the noise tensor to the normal features $F_{\text{clean}}$ for perturbation based on attention mask $A_{\text{final}}$ as follows: 
\begin{equation}
   F' = F_{\text{clean}} + \mathcal{E} \odot (\alpha(t)\cdot\Phi_{\text{norm}}(A_{\text{final}}) +\beta)
\end{equation}
where $\odot$ means the element-wise product, $\alpha(t)$ and $\beta$ control the intensity of adding noise to the features, which linearly increases with the training epochs. $\beta$ is a hyperparameter. $\Phi_{\text{norm}}$ denotes the max-min normalization function. The value of $\alpha(t)$ can be calculated by
\begin{equation}
    \alpha(t) = \gamma \cdot (\frac{t}{T} (p - m)+ m),
\end{equation} 
where $\gamma$ represents the basic noise factor, $p$ denotes the maximum noise intensity, $m$ represents the minimum noise intensity, $T$ represents the maximum training epoch and $t$ represents the current training epoch.

Our key distinction is that the guided attention derives directly from the reconstruction learning process. By directing noise injection into reconstruction-sensitive areas rather than generic salient regions, it guides subsequent training to force the decoder to learn important intrinsic structures.

In addition to introducing noise at the feature level, we also incorporate noise at the image level, guided by the attention mask $A_{\text{final}}$, resulting in improved performance. 
We upsample $A_{\text{final}}$ to $A_{\text{img}} \in \mathbb{R}^{H \times W}$ and progressively increase its binarization mask ratio.
Gaussian noise is also applied within the masked regions. Gradually increasing the mask ratio increases the difficulty of denoising the decoder, which helps model training.

\subsection{Training and Inference}
During training, for each normal image $I_{normal} \in \mathcal{X}_{train}$, we aim to reconstruct its normal features $F_{\text{clean}}$ from the perturbed counterparts at both the image and feature levels. Our total loss is derived as 
\begin{equation}
L_{\text{total}} = \frac{1}{2} (L_{\text{feat}} + L_{\text{img,feat}})
\end{equation}
where $L_{\text{feat}}$ indicates reconstruction by perturbing features, and $L_{\text{img, feat}}$ represents reconstruction by perturbing image and features at the same time, which are defined below:
\begin{equation}
\begin{cases}
\begin{aligned}
L_{\text{feat}} &= \frac{1}{H_f \times W_f} \text{MSE}(F_{\text{feat}}, F_{\text{clean}})\\
L_{\text{img,feat}} &= \frac{1}{H_f \times W_f} \text{MSE}(F_{\text{img,feat}}, F_{\text{clean}})
\end{aligned}
\end{cases}
\end{equation}
where $F_{\text{clean}}$ indicates the clean features of the pretrained encoder, $F_{\text{feat}}$ represents the reconstructed features from perturbed features, and $F_{\text{img,feat}}$  is the reconstructed feature by perturbing images and features at the same time. 

Following~\cite{you2022unified}, we utilize reconstruction errors to calculate the anomaly map during inference. 
We calculate pixel-level reconstruction errors using $L_2$ distance, which are then upsampled to the image size.
For image-level score, we simply use the maximum value.

\section{Experiment}

\subsection{Experimental Setups}
\noindent\textbf{Datasets and Metrics.} We conduct experiments on \emph{MVTec-AD}~\cite{bergmann2019mvtec}, \emph{VisA}~\cite{zou2022spot}, and \emph{MVTec-3D}~\cite{bergmann2021mvtec}. 
Notably, MVTec-AD is evaluated under both one-class and multi-class settings, while for MVTec-3D, we exclusively utilize RGB images. 
Following~\cite{you2022unified}, we adopt the Area Under the Receiver Operating Curve (AUROC) for both image-level detection (I-AUC) and pixel-level localization (P-AUC). Additionally, Per-Region-Overlap (PRO)~\cite{Bergmann_2020} is employed for fine-grained evaluation.

\noindent\textbf{Implementation Details.}
Input images are resized to $224 \times 224$. We adopt a frozen DINO (ViT-S-16)~\cite{caron2021emerging} as the feature extractor. The model is trained for 500 epochs on a single NVIDIA RTX 4090 GPU with a batch size of 32, using AdamW~\cite{loshchilov2019decoupledweightdecayregularization} and a weight decay of $1 \times 10^{-4}$.
The learning rate starts at $1 \times 10^{-3}$ and drops by 0.1 after 200 epochs. Regarding perturbation, the image noise ratio and feature noise intensity $\alpha$ linearly increase from 0.6 to 1.0 (epochs 100 to 400) and 0 to 1.0 (epochs 1 to 400), respectively. For EMA, $\eta$ is set to 0.9999 with updates every 10 steps. Additionally, 32-fold data augmentation is applied for the few-shot setting.
The EMA is simply a weighted sum of the old and new decoders, computed directly without backpropagation, accounting for $<0.2\%$ of the total training time under strict CUDA synchronization. Furthermore, compared to methods relying on pure random noise and complex architectures, AGPNet converges in only 500 epochs (vs. 1000 for UniAD). All efficiency metrics are evaluated at $224 \times 224$ resolution without AMP/TensorRT. We record pure model inference (excluding pre/post-processing) over the entire MVTec-AD dataset, reporting the average FPS across 10 stable runs.

\noindent\textbf{Comparison Protocol.}
The main benchmark results are collected from the corresponding papers or obtained using official implementations, as indicated in the table captions. Since existing methods may employ different backbones, input resolutions, training schedules, and model capacities, these results represent practical benchmark comparisons rather than strictly architecture-controlled comparisons. To isolate the contribution of attention guidance, all variants in Table~\ref{tab:noise_strategies} are evaluated within the same reconstruction framework and training protocol, with only the perturbation strategy changed. We therefore base the attention-specific conclusions primarily on this controlled ablation.

\begin{table}[ht]
\centering
\caption{Comparison of results on MVTec AD, VisA, and MVTec-3D (left: multi-class, right: one-class). $^\ast$: official code results. \textbf{Bold}/\underline{underline}: best/second best.}
\label{tab_fixed_alignment}
\resizebox{\textwidth}{!}{
\begin{tabular}{|l|c|c|c|c|c|c|c|c|c|l|c|c|c|} 
\hline
\multicolumn{10}{|c|}{Multi-class Setting} & \multicolumn{4}{c|}{One-class Setting} \\ 
\hline
\multirow{2}{*}{Method} 
& \multicolumn{3}{c|}{MVTec AD} 
& \multicolumn{3}{c|}{VisA} 
& \multicolumn{3}{c|}{MVTec-3D} 
& \multirow{2}{*}{Method} 
& \multicolumn{3}{c|}{MVTec AD} \\
\cline{2-10} \cline{12-14}
& I-AUC & P-AUC & PRO 
& I-AUC & P-AUC & PRO 
& I-AUC & P-AUC & PRO 
& & I-AUC & P-AUC & PRO \\
\hline
DRÆM*\cite{zavrtanik2021draem} & 88.1 & 87.2 & 71.1 & 79.1 & 91.3 & 59.1 & 65.2 & 93.2 & 55.0 & DRÆM\cite{zavrtanik2021draem} & 98.0 & 97.3 & -- \\
SimpleNet*\cite{liu2023simplenet} & 85.1 & 88.9 & 86.5 & 87.2 & 96.8 & 81.4 & -- & -- & -- & RD4AD~\cite{RD4AD} & 98.5 & 97.8 & \textbf{93.9} \\
UniAD~\cite{you2022unified} & 96.5 & 96.8 & 90.2 & 85.5 & 95.9 & 75.6 & 78.9 & \underline{96.5} & 88.1 & PatchCore\cite{roth2022towards} & 99.1 & \underline{98.1} & \underline{93.5} \\
DiAD\cite{he2024diad} & 97.2 & 96.8 & 90.7 & 86.8 & 96.0 & 75.2 & \underline{84.6} & 96.4 & 87.8 & DeSTSeg\cite{zhang2023destseg} & 98.6 & 97.9 & -- \\
MSTAD\cite{MSTAD} & \underline{98.6} & 97.2 & - & - & - & - & - & - & - & SimpleNet\cite{liu2023simplenet} & \textbf{99.6} & \underline{98.1} & -- \\
MoEAD\cite{meng2024moead} & 97.7 & 97.0 & - & \textbf{93.1} & \textbf{98.7} & - & - & - & - & MSTAD\cite{MSTAD} & 98.3 & 97.5 & -- \\
ViTAD\cite{zhang2025exploring} & 98.3 & \underline{97.7} & \underline{91.4} & 90.5 & 98.2 & \underline{85.1} & 78.7 & \textbf{98.0} & \underline{91.3} &  GLAD\cite{yao2024glad} & \underline{99.3} & \textbf{98.6} & - \\
\rowcolor[HTML]{CFEFFF} \textbf{Ours} & \textbf{98.7} & \textbf{98.0} & \textbf{92.9} & \underline{92.3} & \underline{98.4} & \textbf{85.5} & \textbf{84.9} & \textbf{98.0} & \textbf{92.9} & \textbf{Ours} & 99.2 & \underline{98.3} & 93.1 \\
\hline
\end{tabular}}
\vspace{-2mm}
\end{table}

\begin{figure}[t]
\centering
\includegraphics[width=\textwidth]{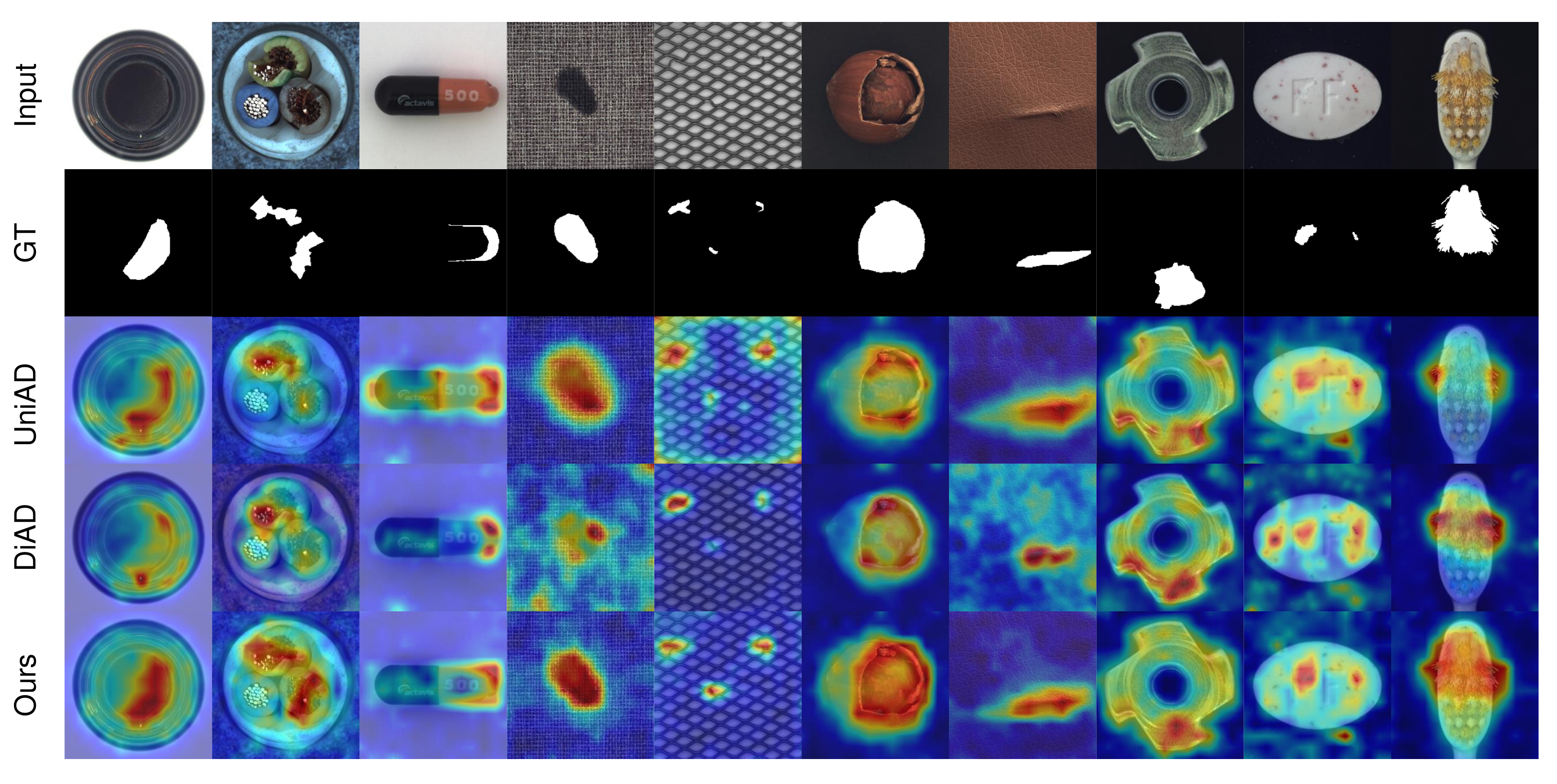}
\caption{Qualitative illustration on MVTec-AD dataset.} \label{fig:compare}
\vspace{-2mm}
\end{figure}

\subsection{Performance on MVTec-AD.}
\noindent\textbf{Evaluation under the multi-class setting.} 
As shown in Table~\ref{tab_fixed_alignment}, it can be observed that applying one-class methods to the multi-class scenario reflects undesirable performance.
Among the methods reported in Table~\ref{tab_fixed_alignment}, AGPNet achieves the best I-AUC of 98.7\%, P-AUC of 98.0\%, and PRO of 92.9\% on MVTec-AD.
On the more challenging PRO metric, our model also attains the top performance of 92.9\%, marking a 1.5\% improvement over ViTAD~\cite{zhang2025exploring}.
To further evaluate the efficiency, Tab.~\ref{tab:perf_eff} compares AGPNet with recent multi-class methods. As shown, our model achieves comparable performance (a 0.5\% drop) despite using less than 10\% learnable parameters, under 6\% FLOPs, and a significantly lower resolution ($224^2$ vs. $392^2$). Similarly, while AF-CLIP~\cite{fang2025af} achieves superior few-shot precision, it relies on a massive CLIP ViT-L at $518^2$ resolution, consuming $\sim294$ GFLOPs ($>50\times$ heavier than ours). These results demonstrate a favorable accuracy--efficiency trade-off for AGPNet.

\begin{table}[ht]
\centering
\caption{Model Performance on MVTec-AD \& Comparison.}
\label{tab:perf_eff}
\resizebox{\textwidth}{!}{
    \begin{tabular}{|l|c|c|c|c|c|} 
        \hline
        Method & P-AUROC(\%) & Resolution & Backbone & \makecell{Learnable \\ Parameters (M)} & FLOPs (G) \\
        \hline
        INP-Former~\cite{INP-Former} & 98.5 & 392 $\times$ 392 & ViT-B (DINOv2) & 53.8 & 98.0 \\
        Dinomaly~\cite{Dinomaly}  & 98.4 & 392 $\times$ 392 & ViT-B (DINOv2) & 46.8 & 104.7 \\
        \textbf{Ours} & 98.0 & 224 $\times$ 224 & ViT-S (DINO) & 4.42 & 5.63 \\
        \hline
    \end{tabular}
}
\vspace{-2mm}
\end{table}

Fig.~\ref{fig:compare} demonstrates the precise anomaly localization of AGPNet on representative samples. 

\noindent\textbf{Evaluation under the one-class setting.} Even in the one-class setting, AGPNet outperforms representative synthetic-based methods (e.g., DRAEM~\cite{zavrtanik2021draem}, DeSTSeg~\cite{zhang2023destseg}) by considerable margins, achieving a competitive P-AUC of 98.3\%.
While SimpleNet~\cite{liu2023simplenet} yields higher detection accuracy in this setting, its performance degrades significantly when adapted to the multi-class scenario.
In contrast, AGPNet maintains competitive one-class performance while showing a clearer advantage in the unified multi-class setting.

\subsection{Anomaly Detection on VisA and MVTec-3D}

\noindent\textbf{Evaluation on VisA.} 
Table~\ref{tab_fixed_alignment} demonstrates our competitive performance compared to existing approaches. 
AGPNet substantially outperforms the pioneering unified framework UniAD~\cite{you2022unified} by significant margins of 6.8\% (I-AUC) and 2.5\% (P-AUC), and consistently exceeds recent strong competitors like ViTAD~\cite{zhang2025exploring} and the diffusion-based DiAD~\cite{he2024diad}.
While MoEAD~\cite{meng2024moead} yields marginally higher scores on this specific dataset, it lacks the cross-dataset consistency demonstrated by AGPNet (e.g., our +1.0\% P-AUC lead on MVTec-AD).

\noindent\textbf{Evaluation on MVTec-3D.}
As presented in Table~\ref{tab_fixed_alignment}, AGPNet achieves the best or tied-best performance across all evaluated metrics.
Compared to UniAD, we observe substantial improvements of 6.0\% and 4.8\% in I-AUC and PRO, respectively.
Furthermore, against the diffusion-based DiAD, AGPNet demonstrates distinct advantages, particularly in localization with a 5.1\% gain in PRO.
We outperform ViTAD in I-AUC and PRO while matching its P-AUC.
These quantitative results substantiate the effectiveness of our proposed framework on this challenging dataset. For MVTec-3D, even with only RGB images, we still attain competitive performance. It fully validates that it can effectively mine key feature information from RGB images and exhibits strong robustness.

\begin{table}
\centering
\caption{Ablation studies on attention-guided noise. `-' means no noise is added. `R' indicates adding random noise. `A' denotes adding attention-guided noise.}
\label{tab:noise_strategies}
\resizebox{\linewidth}{!}{
\begin{tabular}{|l|c|c|c|c|c|c|c|c|c|}
\hline
Noise Type & No Noise & \multicolumn{3}{c|}{Only Random} & \multicolumn{3}{c|}{Only Attention} & \multicolumn{2}{c|}{Hybrid} \\ 
\hline
Image/Feature-level & -/- & R/- & -/R & R/R & A/- & -/A & A/A & A/R & R/A \\
\hline
I-AUC (\%) & 90.5 & 90.7 & 98.0 & 97.7 & 91.9 & 98.3 & \mycolor{\textbf{98.7}} & 98.3 & 98.6 \\
\hline
P-AUC (\%) & 95.5 & 95.5 & 97.7 & 97.8 & 96.0 & 97.7 & \mycolor{\textbf{98.0}} & 97.8 & 97.8 \\
\hline
\end{tabular}}
\vspace{-2mm}
\end{table}

\subsection{Ablation Study}
We mainly analyze AGPNet on MVTec-AD under multi-class setting.

\noindent\textbf{Attention-Guided Noise.}
Table~\ref{tab:noise_strategies} compares different perturbation strategies under the same reconstruction framework. Random feature-level perturbation substantially improves I-AUC from 90.5\% to 98.0\%, showing that perturbation-based training provides the primary gain. Attention guidance further improves the corresponding random variants by 1.2 and 0.3 percentage points in I-AUC at the image and feature levels, respectively, while maintaining the same feature-level P-AUC. Applying attention-guided perturbation at both levels achieves the best results of 98.7\% I-AUC and 98.0\% P-AUC. These results show that attention guidance provides an additional sample-adaptive improvement over generic perturbation.

\noindent\textbf{Compatibility across Backbones.}
We evaluate AGPNet across diverse architectures, including CNNs (Wide-ResNet50~\cite{he2015deepresiduallearningimage}, EfficientNet-b4~\cite{tan2020efficientnetrethinkingmodelscaling}) and ViTs (CLIP~\cite{radford2021learning}, DINO~\cite{caron2021emerging}).
It is worth noting that since CNNs lack the native self-attention maps required for our optimal mask generation, AGPNet operates in a \textit{constrained mode} without full attention guidance.
Despite this architectural limitation, our method still demonstrates robustness, outperforming the CNN-based SimpleNet by 2.3\% (I-AUC) and 4.5\% (P-AUC) using Wide-ResNet50.
When equipped with CLIP, where attention guidance is fully utilized, AGPNet exceeds ViTAD-CLIP by 23.4\% in I-AUC and 14.2\% in P-AUC.
These results suggest that the perturbation framework can be adapted to
different feature extractors, while its full attention-guided formulation
is particularly suitable for attention-based backbones.

\begin{table}
\caption{Ablation studies on backbones and compare with the previous methods in multi-class setting.}
\label{Backbone}
\centering
\small
\setlength{\tabcolsep}{23pt}
\begin{tabular}{|l|l|c|}
\hline
Backbone & Method & I-AUC / P-AUC (\%) \\
\hline
\multirow{2}{*}{WideResNet-50} & SimpleNet & 85.1 / 88.9 \\
\cline{2-3}
& \mycolor{Ours} & \mycolor{87.4 / 93.4} \\ 
\hline
\multirow{2}{*}{EfficientNet-b4} & UniAD & 96.5 / 96.8 \\
\cline{2-3}
& \mycolor{Ours} & \mycolor{97.0 / 96.7} \\ 
\hline
\multirow{2}{*}{CLIP (ViT-B)} & ViTAD-CLIP & 71.2 / 81.6 \\
\cline{2-3}
& \mycolor{Ours} & \mycolor{94.6 / 95.8} \\ 
\hline
\multirow{2}{*}{DINO (ViT-S)} & ViTAD & 98.3 / 97.7 \\
\cline{2-3}
& \mycolor{Ours} & \mycolor{98.7 / 98.0} \\ 
\hline
\end{tabular}
\vspace{-2mm}
\end{table}

\begin{table}
\caption{Ablation Studies on Noise Intensity $\alpha$ at Feature-level and Attention Map. 'P' and 'L' denote utilizing solely $A_{prior}$ and $A_{learn}$ for guidance, respectively, while 'M' represents the combination of both.}
\label{attention map}
\centering
\small
\setlength{\tabcolsep}{10pt}
\begin{tabular}{|l|c|c|c|c|c|c|c|}
\hline
Components & \multicolumn{4}{c|}{Noise Intensity $\alpha$ at Feature-level} & \multicolumn{3}{c|}{Attention map} \\ 
\hline
Options & 0.5 & 1.0 & 1.5 & 0--1.0 & P & L & M \\
\hline
I-AUC (\%) & 98.5 & 98.6 & 98.3 & \mycolor{\textbf{98.7}} & 98.4 & 98.6 & \mycolor{\textbf{98.7}} \\
\hline
P-AUC (\%) & 97.9 & 97.9 & 97.9 & \mycolor{\textbf{98.0}} & 97.9 & 97.9 & \mycolor{\textbf{98.0}} \\ 
\hline
\end{tabular}
\vspace{-2mm}
\end{table}

\noindent\textbf{Easy-to-Hard Noise Addition.}
Recall that in Sec.III, we use a hyperparameter $\alpha$ to control the noise intensity at the feature level. Table~\ref{attention map} explores the impact of noise intensity. The experiments indicate that Easy-to-hard noise with suitable intensity helps the model to model the target distribution and achieve the best performance. Compared with fixed noise of different intensities, the I-AUC and P-AUC are slightly improved (0.1\%-0.4\%).

\noindent\textbf{Attention Map.}
Table~\ref{attention map} analyzes the contribution of attention components.
We observe that while using $A_{\text{prior}}$ or $A_{\text{learn}}$ individually yields satisfactory results, their combination achieves the optimal performance.
Specifically, $A_{\text{prior}}$ provides stable guidance during the early training stages, while $A_{\text{learn}}$ introduces dynamic difficulty to prevent training plateauing in later phases.
Therefore, fusing both ensures a complementary benefit, effectively balancing training stability and difficulty.

\subsection{Extending to Few-shot Anomaly Detection}
To evaluate AGPNet with limited normal training samples, we conduct experiments under the few-shot setting.
As shown in Table~\ref{tab:fewshot-anomaly-detection}, AGPNet achieves a tied-best P-AUC of 96.2\% in the 2-shot setting and the best P-AUC of 97.3\% in the 4-shot setting, demonstrating strong anomaly localization performance. Although PromptAD~\cite{li2024promptad} obtains higher image-level AUROC across the evaluated shot settings, AGPNet remains competitive in image-level detection without relying on a large CLIP backbone or prompt tuning. These results confirm the effectiveness of attention-guided perturbation in capturing intrinsic normal patterns with limited training samples.
Together with the full-data results, these findings indicate that AGPNet remains effective under different amounts of normal training data.

\begin{table}
\caption{Comparison of few-shot image-level/pixel-level AUROC (\%) results on MVTec. \textbf{Bold} and \underline{underline} indicate the best and the second best, respectively.}
\label{tab:fewshot-anomaly-detection}
\centering
\small
\setlength{\tabcolsep}{16.4pt}
\begin{tabular}{|l|c|c|c|}
\hline
\multirow{2}{*}{Method} & 1-shot & 2-shot & 4-shot \\
\cline{2-4}
& \multicolumn{3}{c|}{I-AUC / P-AUC (\%)} \\
\hline
SPADE~\cite{cohen2020sub} & 81.0 / 91.2 & 82.9 / 92.0 & 84.8 / 92.7 \\
\hline
PaDiM~\cite{defard2021padim} & 76.6 / 89.3 & 78.9 / 91.3 & 80.4 / 92.6 \\
\hline
PatchCore~\cite{roth2022towards} & 83.4 / 92.0 & 86.3 / 93.3 & 88.8 / 94.3 \\
\hline
WinCLIP+~\cite{jeong2023winclip} & 93.1 / 95.2 & \underline{94.4} / 96.0 & \underline{95.2} / 96.2 \\
\hline
RWDA~\cite{tamura2023random} & \underline{93.3} / -- & 94.0 / -- & 94.5 / -- \\
\hline
FastRecon~\cite{fang2023fastrecon} & -- / -- & 91.0 / 95.9 & 94.2 / \underline{97.0} \\
\hline
PromptAD~\cite{li2024promptad} & \textbf{94.6} / \textbf{95.9} & \textbf{95.7} / \textbf{96.2} & \textbf{96.6} / 96.5 \\
\hline
\mycolor{\textbf{Ours}} & \mycolor{92.4 / \underline{95.6}} & \mycolor{93.3 / \textbf{96.2}} & \mycolor{\underline{95.2} / \textbf{97.3}} \\
\hline
\end{tabular}
\vspace{-2mm}
\end{table}

\section{Conclusion}
In this paper, we have proposed a simple yet effective reconstruction-based framework AGPNet to alleviate the issue of `identical shortcut' for visual anomaly detection. 
It consists of a branch for reconstruction and an auxiliary branch that aims to generate an attention mask for targeted perturbations. 
To accommodate various samples across diverse categories, the attention mask is based on prior attention weights from the frozen feature extractor and mean-distillation of the decoder. 
Experiments demonstrate competitive performance under the evaluated multi-class and one-class settings, while the few-shot results indicate promising pixel-level localization performance with limited normal samples.

\section{Acknowledgements}
This work is supported by the National Natural Science Foundation of China (No.62302188); Hubei Province Natural Science Foundation (No.2023AFB267); Fundamental Research Funds for the Central Universities (No.2662023XXQD001).

\bibliographystyle{splncs04}
\bibliography{paper}

\end{document}